\documentclass{article} 
\usepackage{proceedings}
\usepackage{latexsym}
\usepackage{amssymb}
\usepackage{amsmath}

\newcommand{\GPL}[2]{#2^{#1\downarrow}}
\newcommand{\LO}[2]{#2^{#1\uparrow}}

\newcommand{\lar}{\leftarrow}
\newcommand{\vph}{\varphi}

\newcommand{\Appx}{\mathit{Appx}}
\newcommand{\At}{\mathit{At}}
\newcommand{\lub}{\mathit{lub}}
\newcommand{\lfp}{\mathit{lfp}}
\newcommand{\glb}{\mathit{glb}}
\newcommand{\leqi}{\leq_p}

\newcommand{\n}{\mbox{\bf not}}
\newcommand{\Tr}{\mbox{\bf t}}
\newcommand{\Fa}{\mbox{\bf f}}

\newcommand{\ra}{\rightarrow}

\newcommand{\KK}{\mathrm{KK}}
\newcommand{\WF}{\mathrm{WF}}

\newcommand{\LC}{L^c}

\newtheorem{theorem}{Theorem}[section]
\newtheorem{proposition}[theorem]{Proposition}

\newtheorem{corollary}[theorem]{Corollary}
\newtheorem{lemma}[theorem]{Lemma}

\title{Ultimate approximations in nonmonotonic knowledge
representation systems} 
 
\author{ {\bf Marc Denecker}\\
Department of Computer Science\\
K.U.Leeuven\\
Celestijnenlaan 200A, B-3001 Heverlee\\
Belgium\\
\And 
{\bf Victor W. Marek}  \\ 
Department of Computer Science\\ 
University of Kentucky\\
Lexington, KY, 40506-0046\\
USA\\
\And 
{\bf Miros\l aw Truszczy\'nski}   \\ 
Department of Computer Science\\
University of Kentucky\\
Lexington, KY, 40506-0046\\
USA\\
} 
 
\begin{document} 
 
\maketitle 
 
\begin{abstract} 
We study fixpoints of operators on lattices. To this end we introduce 
the notion of an approximation of an operator. We order approximations 
by means of a {\em precision} ordering. We show that each 
lattice operator $O$ has a unique most precise or {\em ultimate} 
approximation. We demonstrate that fixpoints of this ultimate 
approximation provide useful insights into fixpoints of the 
operator $O$.

We apply our theory to logic programming and introduce the ultimate
Kripke-Kleene, well-founded and stable semantics. We show that 
the ultimate Kripke-Kleene and well-founded semantics are 
more precise then their standard counterparts We argue that 
ultimate semantics for logic programming have attractive epistemological 
properties and that, while in general they are computationally more 
complex than the standard semantics, for many classes of theories, their
complexity is no worse.

\end{abstract} 
 
\section{INTRODUCTION}\label{intro}

Semantics of most knowledge representation languages are defined as
collections of interpretations or possible-world structures.  The sets
of interpretations and possible-world structures, with some natural
orderings, form complete lattices. Logic programs, and default and
autoepistemic theories determine operators on these lattices. In many
cases, semantics of programs and theories are given as fixpoints of
these operators. Consequently, an abstract framework of lattices,
operators on lattices and their fixpoints has emerged as a powerful
tool in investigations of semantics of these logics. 
Studying semantics of nonmonotonic reasoning systems within an
algebraic framework allows us to eliminate inessential details
specific to a particular logic, simplify arguments and find common
principles underlying different nonmonotonic formalisms.

The roots of this algebraic approach can be traced back to studies of
semantics of logic programs \cite{vEK76,ave82,fi85,przy90} and of
applications of lattices and bilattices in knowledge representation
\cite{gin88}. Exploiting the concept of a bilattice and relying on
some general properties of operators on lattices and bilattices,
Fitting proposed an elegant algebraic treatment of all major 2-, 3-
and 4-valued semantics of logic programs \cite{fi99}, that is, the
supported-model semantics \cite{cl78}, stable-model semantics
\cite{gl88}, Kripke-Kleene semantics \cite{fi85,kun87} and
well-founded semantics \cite{vrs91}.

In \cite{dmt00a}, we extended Fitting's work to a more abstract
setting of the study of fixpoints of lattice operators.  Central to
our approach is the concept of an approximation of a lattice operator
$O$. An approximation is an operator defined on a certain bilattice
(the product of the lattice by itself, with two appropriately defined
lattice orderings). Using purely algebraic techniques, for an
approximation operator for $O$ we introduced the notion of the stable
operator and the concepts of the Kripke-Kleene, well-founded and
stable fixpoints, and showed how they provide information about
fixpoints of the operator $O$. In \cite{dmt00a} we noted that our
approach generalizes the results described in \cite{fi99}. We observed
that the 4-valued immediate consequence operator ${\cal T}_P$ is an
approximation operator for the 2-valued immediate consequence operator
$T_P$ and showed that all the semantics considered by Fitting can be
derived from ${\cal T}_P$ by means of the general algebraic
constructions that apply to {\em arbitrary} approximation operators.

In \cite{dmt00}, we applied our algebraic approach to default and
autoepistemic logics. Autoepistemic logic was defined by Moore \cite{mo84} 
to formalize the knowledge of a rational agent with full introspection
capabilities. 
In Moore's approach, an autoepistemic theory $T$ defines a characteristic 
operator $D_T$ on the lattice of all possible-world structures. Fixpoints
of $D_T$ (or, to be precise, their
theories) are known as {\em expansions}. In \cite{dmt00}, we proposed 
for $D_T$ an approximation operator, ${\cal D}_T$, defined on a 
bilattice of belief pairs (pairs of possible-world structures). Complete 
fixpoints of ${\cal D}_T$ correspond to expansions of $T$ (fixpoints of 
$D_T$), the least fixpoint of ${\cal D}_T$ provides a constructive 
approximation to all expansions (by analogy with logic programming, we
called it the {\em Kripke-Kleene fixpoint}). Using general techniques
introduced in \cite{dmt00a} we derived from ${\cal D}_T$ its {\em stable}
counterpart, the operator ${\cal D}_T^{st}$. Complete fixpoints of 
${\cal D}_T^{st}$ yield a new semantics of {\em extensions} for 
autoepistemic logic. Finally, the least fixpoint of the stable 
operator results in yet another new semantics, the well-founded 
semantics for autoepistemic logic (again, called so due to analogies 
to the well-founded semantics in logic programming), which approximates 
all extensions.

The same picture emerged in the case of default logic
\cite{dmt00}. For a default theory $\Delta$ we defined an operator
${\cal E}_\Delta$ and characterized all major semantics for default
logic in terms of fixpoints of ${\cal E}_\Delta$. In particular, the
standard semantics of extensions \cite{re80} is determined by complete
fixpoints of the stable operator ${\cal E}^{st}_\Delta$ derived from
${\cal E}_\Delta$.  Our results on autoepistemic and default logics
obtained in \cite{dmt00} allowed us to clarify the issue of their
mutual relationship and provided insights into fundamental constructive
principles underlying these two modes of nonmonotonic reasoning.

These result prove that the algebraic framework developed in \cite{dmt00a} is
an effective tool in studies of semantics of knowledge representation
formalisms. It allowed us to establish a comprehensive semantic
treatment for nonmonotonic logics and demonstrated that major
nonmonotonic systems are closely related. However, the approach, as it
was developed, is not entirely satisfactory. It provides no
criteria that would allow us to prefer one approximation over another
when attempting to define the concept of a stable fixpoint or when
approximating fixpoints by means of the Kripke-Kleene or well-founded
fixpoints.  It does not give us any general indications how to obtain
approximations and which
approximation to pick. Thus, our theory leaves out a key link in the
process of defining and approximating fixpoints of operators on
lattices.

In particular, when defining semantics of nonmonotonic formalisms, we 
{\em select} an approximation operator, rather then derive it in a
principled way. The approximations used, the bilattice operators ${\cal 
T}_P$, ${\cal D}_T$ and ${\cal E}_\Delta$, are not algebraically 
determined by their corresponding lattice operators $T_P$, $D_T$ and 
$E_\Delta$, respectively. Consequently, some programs or
theories with the same basic operators have different
Kripke-Kleene, well-founded or stable fixpoints associated with them.

We address this problem here. We extend our theory of approximations 
and introduce the notion of the precision of an approximation. We show 
that each lattice operator $O$ has a unique {\em most precise} 
approximation which we call the {\em ultimate approximation} of
$O$. Since the ultimate approximation is determined by $O$, it is 
well suited for investigations of fixpoints of $O$. As a result we 
obtain concepts of ultimate stable fixpoints, the ultimate Kripke-Kleene 
fixpoint and the ultimate well-founded fixpoint that depend on $O$ only 
and not on a (possibly arbitrarily) selected approximation to $O$. 

We apply our theory to logic programming, default logic and
autoepistemic logic (only the first system is discussed here,
due to space limitations). We compare ultimate 
semantics with the corresponding ``standard'' semantics of logic 
programs. In particular, we show that the ultimate Kripke-Kleene and 
the ultimate well-founded semantics are more precise then the standard 
Kripke-Kleene and well-founded semantics. This better accuracy comes, 
however, at a cost. We show that ultimate semantics are in general 
computationally more complex. On the other hand, we show that for wide 
classes of theories, including theories likely to occur in practice, 
the complexity remains the same. Thus, our new semantics may prove 
useful in computing stable models and default extensions.

The ultimate semantics have also properties that are attractive from 
the logic perspective. In particular, two programs or theories determining
the same basic 2-valued operator have the same ultimate semantics.
This property, as we noted, is not true in the standard case.

In summary, our contributions are as follows. We extend the algebraic
theory of approximations by providing a principled way of deriving an
approximation to a lattice operator. In this way, we obtain concepts of 
Kripke-Kleene fixpoint, well-founded fixpoint and stable fixpoints that
are determined by the operator $O$ and not by the choice of an
approximation. In specific contexts of most commonly used nonmonotonic
systems we obtain new semantics with desirable logical properties and
possible computational applications. 

\section{PRELIMINARIES}
\label{prel}

Let $\langle L,\leq \rangle$ be a poset and let $A$ be an operator on
$L$. A poset is {\em chain-complete} if it contains the least element
$\bot$ and if every chain of elements of $L$ has a least upper bound
($\lub$) in $L$. An element $x$ of $A$ is a {\em pre-fixpoint} of $A$
if $A(x)\leq x$; $x$ is a {\em fixpoint} of $A$ if $A(x)=x$.

Let $A$ be a monotone operator on a chain-complete poset $\langle
L,\leq \rangle$. Let us define a sequence of elements of $L$ by
transfinite induction as follows: (1) $c^0 = \bot$; (2)
$c^{\alpha+1} = A(c^{\alpha})$; 
(3) $c^{\alpha} = \lub(\{c^{\beta}\colon \beta < \alpha\})$, for a
limit ordinal $\alpha$.  One can show that this sequence is well
defined, that is has in $L$ its least upper bound and that this
least upper bound is the least fixpoint of $A$ ($\lfp(A)$, in
symbols). One can also show that the least fixpoint of a monotone
operator on a chain-complete poset is the least pre-fixpoint of
$A$. That is, we have $\lfp(A) = \glb(\{x\in L\colon A(x) \leq x\})$.
Monotone operators on chain-complete posets and their fixpoints and
pre-fixpoints are discussed in \cite{mrk76}.

A lattice is a poset $\langle L, \leq \rangle$ such that
$L\not=\emptyset$ and every pair of elements $x,y \in L$ has 
a unique greatest lower bound and least upper bound. A lattice is 
{\em complete} if its {\em every} subset has a greatest lower
bound and a least upper bound. In particular, a complete lattice has a
least and a greatest element denoted by $\bot$ and $\top$, 
respectively.

For any two elements $x,y\in L$, we define $[x,y]=\{z\in L\colon x\leq
z\leq y\}$. If $\langle L,\leq \rangle$ is a complete lattice and
$x\leq y$, then $\langle [x,y],\leq\rangle$ is a complete lattice, too.

Let $\langle L, \leq \rangle$ be a complete lattice. By the {\em
product bilattice} \cite{gin88} of $\langle L,\leq\rangle$ we mean the set 
$L^2=L\times L$ with the following two orderings $\leqi$ and $\leq$:
\newcounter{ct33}
\begin{list}{\arabic{ct33}.\ }{\usecounter{ct33}\topsep 0.03in
\parsep 0in\itemsep 0in\labelsep 0in}
\item $(x,y)\leqi (x',y')\ \ \ \mbox{if}\ \ x\leq x'\ \mbox{and}\ y'\leq
y$
\item $(x,y)\leq (x',y')\ \ \ \mbox{if}\ \ x\leq x'\ \mbox{and}\ y\leq
y'$.
\end{list}
Both orderings are complete lattice orderings for $L^2$. However, in this
paper we are mostly concerned with the ordering $\leqi$.

An element $(x,y)\in L^2$ is {\em consistent} if $x\leq y$. We can think
of a consistent element $(x,y)\in L^2$ as an {\em approximation} to every 
$z\in L$ such that $x \leq z \leq y$. With this interpretation in mind, the 
ordering $\leqi$, when restricted to consistent elements, can be viewed 
as a {\em precision} ordering. Consistent pairs that are ``higher'' in the 
ordering $\leqi$ provide tighter approximations. Maximal consistent 
elements with respect to $\leqi$ are pairs of the form $(x,x)$.
We call approximations of the form $(x,x)$ --- {\em exact}.

We denote the set of all consistent pairs in $L^2$ by $\LC$. The set 
$\langle L^c ,\leq_p\rangle$ is not a lattice. It is, however, 
chain\-complete. Indeed, the element $(\bot,\top)$ is the least element 
in $\LC$ and the following result shows that every chain in $\LC$ has 
(in $\LC$) the least upper bound. 

\begin{proposition}
\label{chain}
Let $L$ be a complete lattice. If $\{(a^\alpha,b^\alpha)\}_\alpha$ is
a chain of elements in $\langle\LC ,\leq_p \rangle$ 
then $\lub(\{a^\alpha\}_\alpha) \leq 
\glb(\{a^\alpha\}_\alpha)$ and $(lub(\{a^\alpha\}_\alpha), 
\glb(\{a^\alpha\}_\alpha)) = \lub_{\leqi}(\{(a^\alpha, 
b^\alpha)\}_\alpha)$. 
\end{proposition}

It follows that every $\leqi$-monotone operator on $\LC$
has a least fixpoint. 

\section{PARTIAL APPROXIMATIONS}
\label{pa}

For an operator $A:\LC \ra\LC$, we denote by $A^1$ and $A^2$ its
projections to the first and second coordinates, respectively. Thus, 
for every $(x,y)\in \LC$, we have $A(x,y)=(A^1(x,y),A^2(x,y))$. An 
operator $A:\LC \ra\LC$ is a {\em partial approximation} operator
if it is $\leqi$-monotone and if for every $x \in L$, $A^1(x,x)=
A^2(x,x)$. We denote the set of all partial approximation operators 
on $\LC$ by $\Appx(\LC)$. Let $A\in \Appx(\LC)$. Since $A$ is
$\leqi$-monotone and $\LC$ is chain-complete, $A$ has a least fixpoint, 
called the {\em Kripke-Kleene fixpoint} of $A$ ($\KK(A)$, in symbols). 
Directly from the definition, it follows that $\KK(A)$
approximates all fixpoints of $A$.

If $A\in \Appx(\LC)$ and $O\colon L\ra L$ is an operator on $L$ such 
that $A(x,x)=(O(x),O(x))$ then we say that $A$ is a {\em partial
approximation of $O$}. We denote the set of all partial
approximations of $O$ by $\Appx(O)$. If $A$ is a partial approximation
of $O$ then $x\in L$ is a fixpoint of $O$ if and only if $(x,x)$ is a 
fixpoint of $A$. Thus, for every fixpoint $x$ of $O$, we have $\KK(A)
\leqi (x,x)$ or, equivalently, $\KK^1(A)\leq x\leq \KK^2(A)$, where 
$\KK^1(A)$ and $\KK^2(A)$ are the two components of the pair $\KK(A)$.

Operators from $\Appx(\LC)$ describe ways to revise consistent
approximations. Of particular interest are those situations when
the revision of an approximation leads to another one that is at 
least as accurate. Let $A$ be an operator on $\LC$. We call an 
approximation $(a,b)$ {\em $A$-reliable} if $(a,b)\leqi A(a,b)$.

\begin{proposition}
\label{rel1}
Let $L$ be a complete lattice and $A\in \Appx(\LC)$.
If $(a,b)\in L^c$ is $A$-reliable then, for every $x\in
[\bot,b]$, $A^1(x,b)\in [\bot,b]$ and, for every $x\in
[a,\top]$, $A^2(a,x)\in [a,\top]$.
\end{proposition}
Proof: Let $x\in [\bot,b]$. Then $(x,b)\leqi(b,b)$. By
the $\leqi$-monotonicity of $A$,
\[
A^1(x,b)\leq A^1(b,b)=A^2(b,b)\leq A^2(a,b) \leq b.
\]
The last inequality follows from the fact that $(a,b)$ is $A$-reliable.
The second part of the assertion can be proved in a similar manner.
\hfill$\Box$

This proposition implies that for every $A$-reliable pair $(a,b)$, the
restrictions of $A^1(\cdot,b)$ to $[\bot,b]$ and $A^2(a,\cdot)$ to
$[a,\top]$ are in fact operators on $[\bot,b]$ and $[a,\top]$,
respectively. Moreover, they are $\leq$-monotone operators on the
posets $\langle[\bot,b],\leq\rangle$ and
$\langle[a,\top],\leq\rangle$. Since $\langle[\bot,b],\leq\rangle$ and
$\langle[a,\top],\leq\rangle$ are complete lattices, the operators
$A^1(\cdot,b)$ and $A^2(a,\cdot)$ have least fixpoints in the lattices
$\langle[\bot,b], \leq\rangle$ and $\langle[a,\top],\leq\rangle$,
respectively. We define:
\[
\GPL{A}{b}=\lfp(A^1(\cdot,b))\ \ \ \mbox{and}\ \ \ \LO{A}{a}
=\lfp(A^2(a,\cdot)).
\]
We call the mapping $(a,b)\mapsto (\GPL{A}{b},\LO{A}{a})$, defined on 
the set of $A$-reliable elements of $\LC$, the {\em stable revision
operator for $A$}. When $A$ is clear from the context, we will drop the
reference to $A$ from the notation.

Directly from the definition of the stable revision operator it follows
that for every $A$-reliable pair, $\GPL{}{b}\leq b$ and $a\leq \LO{}{a}$. 

The stable revision operator for $A\in \Appx(\LC)$ is crucial.
It allows us to distinguish an important subclass of the 
class of all fixpoints of $A$. Let $L$ be a complete lattice and let
$A\in \Appx(\LC)$. We say that $(x,y)\in \LC$ is a {\em stable fixpoint} 
of $A$ if $(x,y)$ is $A$-reliable and is a fixpoint of the stable
revision operator (that is, $x=\GPL{}{y}$ and $y=\LO{}{x}$).
By the $A$-reliability of $(x,y)$, the second requirement is well defined. 

Stable fixpoints of an operator are, in particular, its fixpoints. 

\begin{proposition}\label{rel3}
Let $L$ be a complete lattice and let $A\in \Appx(\LC)$. If $(x,y)$ is 
a stable fixpoint of $A$ then $(x,y)$ is a fixpoint of $A$.
\end{proposition}
Proof: Since $(x,y)$ is stable, $x=\lfp(A^1(\cdot,y))$. In particular, 
$x=A^1(x,y)$. Similarly, $y=A^2(x,y)$. \hfill$\Box$

Let $O$ be an operator on a complete lattice $L$ and let $A\in
\Appx(O)$. We say that $x$ is an {\em $A$-stable} fixpoint of $O$ if
$(x,x)$ is a stable fixpoint of $A$. The notation is justified. Indeed,
it follows from Proposition \ref{rel3} and our earlier remarks that 
every stable fixpoint of $O$ is, in particular, a fixpoint of $O$.

The notion of $A$-reliability is not strong enough to guarantee
desirable properties of the stable revision operator. In particular, 
if $(a,b)\in \LC$ is $A$-reliable, it is not true in general that
$(\GPL{}{b},\LO{}{a})$ is consistent nor that $(a,b)\leqi
(\GPL{}{b},\LO{}{a})$. There is, however, a class of $A$-reliable
pairs for which both properties hold. An $A$-reliable approximation
$(a,b)$ is {\em $A$-prudent} if $a \leq \GPL{}{b}$. We note that
every stable fixpoint of $A$ is $A$-prudent. We will now prove
several basic properties of $A$-prudent approximations.

\begin{proposition}
\label{pru1}
Let $L$ be a complete lattice, $A\in \Appx(\LC)$ and
$(a,b)\in \LC$ be $A$-prudent. Then, $(\GPL{}{b},\LO{}{a})$ 
is consistent, $A$-reliable and $A$-prudent and $(a,b)\leqi 
(\GPL{}{b},\LO{}{a})$.
\end{proposition}
Proof: By the definition of $\GPL{}{b}$ and $\LO{}{a}$ we have that
$\GPL{}{b}\leq b$ and $a\leq \LO{}{a}$. Moreover, since $(a,b)$ is
$A$-prudent, it follows that $a \leq \GPL{}{b}$.

Next, since $(a,b)$ is $A$-reliable, it follows that $a\leq b$ and
$A^2(a,b)\leq b$. Thus, $b$ is a pre-fixpoint of $A^2(a,\cdot)$.
Consequently, $\LO{}{a}\leq b$ (as $\LO{}{a}$ is the least fixpoint of
$A^2(a,\cdot)$). Hence, $(a,b)\leqi(\GPL{}{b},\LO{}{a})$.

By the $\leqi$-monotonicity of $A$ we obtain:
\[
A^1(\LO{}{a},b) \leq A^1(\LO{}{a},\LO{}{a})= A^2(\LO{}{a},\LO{}{a})
\leq A^2(a,\LO{}{a}) = \LO{}{a}.
\]
It follows that $\LO{}{a}$ is a pre-fixpoint of the operator
$A^1(\cdot,b)$.  Thus, $\GPL{}{b}=\lfp(A^1(\cdot,b)) \leq \LO{}{a}$
and so, $(\GPL{}{b},\LO{}{a})$ is consistent.

Let us now observe that $\GPL{}{b} = A^1(\GPL{}{b},b) \leq
A^1(\GPL{}{b},\LO{}{a })$. Similarly, $\LO{}{a} = A^2(a,\LO{}{a}) \geq
A^2(\GPL{}{b},\LO{}{a})$. Thus, the pair $(\GPL {}{b},\LO{}{a})$ is
reliable.

Lastly, we note that for every $x\in [\bot,\LO{}{a}]$, $A^1(x,b) \leq
A^1(x,\LO{}{a})\leq \LO{}{a}$ (the last inequality follows by the
$A$-reliability of $(\GPL{}{b},\LO{}{a})$).  Hence,
$\GPL{}{b}=\lfp(A^1(\cdot,b))\leq \lfp(A^1(\cdot,\LO{}{a}))$ and,
consequently, $(\GPL {}{b},\LO{}{a})$ is $A$-prudent. \mbox{\ \ \ \ }\hfill$\Box$

Let us observe that an $A$-reliable pair $(a,b)$ is revised by 
an operator $A$ into a
more accurate approximation $A(a,b)$. An $A$-prudent pair $(a,b)$
can be revised ``even more''. Namely, it is easy to see that
$A^1(a,b)\leq A^1(\GPL{}{b},b)=\GPL{}{b}$ and
$\LO{}{a}=A^2(a,\LO{}{a})\leq A^2(a,b)$. Thus, $A(a,b)\leqi 
(\GPL{}{b},\LO{}{a})$. In other words, $(\GPL{}{b},\LO{}{a})$
is indeed at least as precise revision of $(a,b)$ as $A(a,b)$ is.

The stable revision operator satisfies a certain monotonicity 
property.

\begin{proposition}\label{Prop.monotonicity}
Let $L$ be a complete lattice and let $A\in \Appx(\LC)$. If
$(a,b)\in \LC$ is $A$-reliable, $(c,d)\in\LC$ is $A$-prudent and if
$(a,b)\leqi (c,d)$, then $(\GPL{}{b},\LO{}{a})\leqi
(\GPL{}{d},\LO{}{c})$.
\end{proposition}
Proof: Clearly, we have $\GPL{}{d}\leq \LO{}{c}\leq d \leq b$. By the
$\leqi$-monotonicity of $A$, it follows that
$A^1(\GPL{}{d},b)\leq A^1(\GPL{}{d},d)=\GPL{}{d}$. Thus,
$\GPL{}{d}$ is a pre-fixpoint of $A^1(\cdot,b)$. Since $\GPL{}{b}$ is
the least fixpoint of $\lfp(A^1(\cdot,b))$, it follows that 
$\GPL{}{b}\leq \GPL{}{d}$.

It remains to prove that $\LO{}{c}\leq \LO{}{a}$.
Let $u= \glb(\LO{}{a},\GPL{}{d})$. By Proposition \ref{pru1},
$(c,d) \leqi (\GPL{}{d},\LO{}{c})$. Since $(a,b)\leqi (c,d)$, it
follows that $a \leq \GPL{}{d}$. Further, by the $A$-reliability of
$(a,b)$ and $(c,d)$, we have $a\leq \LO{}{a}$ and $\GPL{}{d}\leq d$. 
Thus, $a\leq u \leq \LO{}{a}$ and $u\leq \GPL{}{d}\leq d$. Consequently,
\[
A^1(u,d) \leq A^1(u,u) = A^2(u,u) \leq A^2(a,\LO{}{a})=\LO{}{a}
\]
and
\[
A^1(u,d) \leq A^1(\GPL{}{d},d) =\GPL{}{d}.
\]
It follows that $A^1(u,d)\leq \glb(\LO{}{a},\GPL{}{d})=u$. In
particular, $u$ is a pre-fixpoint of $A^1(\cdot,d)$. Since $\GPL{}{d}$ 
is the least fixpoint of $A^1(\cdot, d)$, $\GPL{}{d} \leq u$. Hence, 
$\GPL{}{d} \leq \LO{}{a}$. 

We now have $a\leq c\leq \GPL{}{d} \leq \LO{}{a}$ (the first
inequality follows from the assumption $(a,b)\leq (c,d)$, the second
one follows by Proposition \ref{pru1} from the assumption that $(c,d)$
is $A$-prudent). Thus, $a\leq c\leq \LO{}{a}$ and the 
$\leqi$-monotonicity of $A$ implies
\[
A^2(c,\LO{}{a})\leq A^2(a,\LO{}{a}) = \LO{}{a}.
\]
Hence, $\LO{}{a}$ is a pre-fixpoint of $A^2(c,\cdot)$. Since
$\LO{}{c}$ is the least fixpoint of $A^2(c,\cdot)$, it follows that 
$\LO{}{c}\leq \LO{}{a}$. \hfill$\Box$

Since stable fixpoints are prudent, we obtain the following corollary.

\begin{corollary}
\label{pru3}
Let $L$ be a complete lattice, $A\in \Appx(\LC)$ and let
$(c,d)\in \LC$ be a stable fixpoint of $A$. If $(a,b)\in \LC$ is
$A$-reliable and $(a,b)\leqi (c,d)$ then $(\GPL{}{b},\LO{}{a})\leqi 
(c,d)$. \hfill$\Box$
\end{corollary}

The next result states that the limit of a chain of $A$-prudent pairs 
is $A$-prudent. 

\begin{proposition}
\label{pru2}
Let $L$ be a complete lattice, $A\in \Appx(\LC)$ and
let $\{(a^\alpha,b^\alpha)\}_\alpha$ be a chain of $A$-prudent pairs
from $\LC$. Then, $\lub(\{(a^\alpha,b^\alpha)\}_\alpha)$ is $A$-prudent.
\end{proposition}
Proof: Let us set $a^\infty = \lub(\{a^\alpha\}_\alpha)$ and 
$b^\infty=\glb(\{b^\alpha\}_\alpha)$. By Proposition \ref{chain}, 
$(a^\infty,b^\infty)$ is consistent and $(a^\infty,b^\infty) = 
\lub(\{(a^\alpha,b^\alpha)\}_\alpha)$. Let us now observe that, by 
$A$-reliability of $(a^\alpha,b^\alpha)$ and $\leqi$-monotonicity of 
$A$, we have 
$(a^\alpha,b^\alpha)\leqi A(a^\alpha,b^\alpha) \leqi
A(a^\infty,b^\infty).$
Thus, 
$(a^\infty,b^\infty)= \lub(\{(a^\alpha,b^\alpha)\}_\alpha) \leq
A(a^\infty,b^\infty).$
It follows that $(a^\infty,b^\infty)$ is $A$-reliable. 

The $A$-reliability of $(a^\infty,b^\infty)$ implies, in particular,
that for every $x\in[\bot,b^\infty]$, $A^1(x,b^\infty) \leq
b^\infty$. Thus, by $\leqi$-monotonicity of $A$, for every 
$x\in[\bot,b^\infty]$ 
\[
A^1(x,b^\alpha) \leq A^1(x,b^\infty) \leq b^\infty .
\]
Hence, pre-fixpoints of $A^1(\cdot,b^\infty)$ are prefixpoints
of $A^1(\cdot,b^\alpha)$ and, consequently,
\[
\lfp(A^1(\cdot,b^\alpha)) \leq \lfp(A^1(\cdot,b^\infty)).
\]
Since $(a^\alpha,b^\alpha)$ is $A$-prudent, we have that $a^\alpha\leq
\lfp(A^1(\cdot,b^\alpha))$. Thus, for arbitrary $\alpha$,
$a^\alpha\leq \lfp(A^1(\cdot,b^\infty))$ and, consequently,
$a^\infty\leq\lfp(A^1(\cdot,b^\infty))$. It follows that $(a^\infty,
b^\infty)$ is $A$-prudent. \hfill$\Box$

We will now prove that the set of all stable fixpoints of an operator
has a least element (in particular, it is not empty).
To this end, we define a sequence $\{(a^{\alpha},b^{\alpha})\}_{\alpha}$ 
of elements of $\LC$ by transfinite induction:
\newcounter{ct34}
\begin{list}{\arabic{ct34}.\ }{\usecounter{ct34}\topsep 0.03in
\parsep 0in\itemsep 0in\labelsep 0in}
\item $(a^0,b^0) = (\bot,\top)$
\item If $\alpha=\beta+1$, we define $a^\alpha = \GPL{}{{b^\beta}}$ and 
$b^\alpha = \LO{}{{a^\beta}}$
\item  If $\alpha$ is a limit ordinal, we define $(a^{\alpha}, b^\alpha) = 
\lub(\{(a^{\beta},b^\beta) \colon \beta < \alpha\})$.
\end{list}

\begin{theorem}
\label{wfs}
The sequence $\{(a^{\alpha},b^{\alpha})\}_{\alpha}$ is well defined,
$\leqi$-monotone and its limit is the least stable fixpoint
of a partial approximation operator $A$.
\end{theorem}
Proof: It is obvious that $(\bot,\top)$ is $A$-prudent. Thus, by the
transfinite induction it follows that each element in the sequence is
well defined and $A$-prudent (Propositions \ref{pru1} and \ref{pru2} 
settle the cases of successor ordinals and limit ordinals, respectively). 
In the same way, one can establish the $\leqi$-monotonicity of the 
sequence. 

Let $(a^\infty,b^\infty) = \lub(\{(a^{\beta},b^\beta)\}_\alpha)$.  By
Proposition \ref{pru2}, $(a^\infty,b^\infty)$ is $A$-prudent. Thus,
$(a^\infty,b^\infty)$ is $A$-reliable. Moreover, we have $a^\infty =
\GPL{}{(b^\infty)}$ and $b^\infty=\LO{}{(a^\infty)}$.  Thus,
$(a^\infty,b^\infty)$ is a stable fixpoint of $A$. Further, it is easy
to see by transfinite induction and 
Corollary \ref{pru3} that $(a^\infty,b^\infty)$ approximates all
stable fixpoints of $A$.  Thus, it is the least stable fixpoint of
$A$. \hfill$\Box$

We call this least stable fixpoint the {\em well-founded fixpoint} of 
$A$ and denote 
it by $\WF(A)$. The well-founded fixpoint approximates all stable
fixpoints of $A$. In particular, it approximates all $A$-stable
fixpoints of the operator $O$. That is, for every $A$-stable fixpoint
$x$ of $O$, $\WF(A)\leqi(x,x)$ or, equivalently, $\WF^1(A)\leq
x\leq \WF^2(A)$, where $\WF^1(A)$ and $\WF^2(A)$ are the two components
of the pair $\WF(A)$. Moreover, the well-founded fixpoint is more precise 
than the Kripke-Kleene fixpoint: for $A\in Appx(O)$, $\KK(A)\leqi\WF(A)$. 

In \cite{dmt00,dmt00a}, we showed that when applied to appropriately
chosen approximation operators in logic programming, default logic and
autoepistemic logic, these algebraic concepts of fixpoints, stable
fixpoints, the Kripke-Kleene fixpoint and the well-founded fixpoint   
provide all major semantics for these nonmonotonic systems and allow us
to understand their interrelations.

We need to emphasize that the concept of a partial approximation 
introduced here is different from the concept of {\em approximation} 
introduced in \cite{dmt00a}. The latter notion is defined as an 
operator of the whole bilattice $L^2$. That choice was motivated by 
our search for generality and potential applications of inconsistent 
fixpoints in situations when we admit a possibility of some statements
being overdefined. While different, both approaches are very closely 
related\footnote{We will include a detailed discussion of the 
relationship between the two approaches in the full version of the 
paper.}. 

\section{ULTIMATE APPROXIMATIONS}

Partial approximations in $\Appx(\LC)$ can be ordered. Let $A,B\in
\Appx(\LC)$. We say that $A$ is {\em less precise} than $B$ ($A\leqi
B$, in symbols) if for each pair $(x,y) \in \LC$, $A(x,y) \leqi
B(x,y)$. It is easy to see that if $A\leqi B$ then there is an 
operator $O$ on the lattice $L$ such that $A,B\in \Appx(O)$.

\begin{lemma} \label{Lem.more.precise.revision}
Let $L$ be a complete lattice and $A,B\in \Appx(\LC)$.
If $A\leqi B$ and $(a,b)\in \LC$ is $A$-prudent then $(a,b)$ is
$B$-prudent and $(\GPL{A}{b},\LO{A}{a})\leqi (\GPL{B}{b},\LO{B}{a})$.
\end{lemma}
Proof: Clearly, $(a,b) \leqi A(a,b) \leq B(a,b)$. Thus, $(a,b)$ is 
$B$-reliable. 

For each pre-fixpoint $x\leq b$ of $B^1(\cdot,b)$, 
$A^1(x,b) \leq B^1(x,b) \leq x$. Consequently, $x$ is a prefixpoint of
$A^1(\cdot,b)$.  It follows that $\GPL{A}{b} \leq \GPL{B}{b}$. Since
$a\leq \GPL{A}{b}$, $a\leq \GPL{B}{b}$. Thus $(a,b)$ is $B$-prudent.

Likewise, we can prove that any pre-fixpoint of $A^2(a,\cdot)$ is a
prefixpoint of $B^2(a,\cdot)$, and consequently,
$\LO{B}{a}\leq \LO{A}{a}$. Since also $\GPL{A}{b} \leq \GPL{B}{b}$, it 
follows that $(\GPL{A}{b},\LO{A}{a})\leqi (\GPL{B}{b},\LO{B}{a})$.  
\hfill$\Box$
\medskip

More precise approximation have more precise Kripke-Kleene and
well-founded fixpoints.

\begin{theorem}\label{TKK}
Let $O$ be an operator on a complete lattice $L$. Let $A, B \in 
\Appx(O)$. If  $A\leqi B$ then $\KK(A) \leqi \KK(B)$ and 
$\WF(A)\leqi \WF(B)$. 
\end{theorem}
Proof: Let us denote by $\{(a_A^\alpha,b_A^\alpha)\}_{\alpha}$ the
sequence of elements of $\langle L^c.\leqi\rangle$ obtained by iterating the
operator $A$ over $(\bot,\top)$. The sequence
$\{(a_B^\alpha,b_B^\alpha)\}_{\alpha}$ is defined in the same way.
Since $A\leqi B$, it follows by an easy induction that for every
ordinal $\alpha$, $(a_A^\alpha,b_A^\alpha) \leqi
(a_B^\alpha,b_B^\alpha)$. Since $\KK(A)$ is the limit of the sequence
$\{(a_A^\alpha,b_A^\alpha)\}_{\alpha}$ and $\KK(B)$ is the limit of the
sequence $\{(a_B^\alpha,b_B^\alpha)\}_{\alpha}$, it follows that
$\KK(A) \leqi \KK(B)$.

To prove the second part of the assertion, we will now assume that
the sequences $\{(a_A^\alpha,b_A^\alpha)\}_{\alpha}$ and
$\{(a_B^\alpha,b_B^\alpha)\}_{\alpha}$ denote the sequences used in the
definition of the well-founded fixpoints of $A$ and $B$, respectively.
To prove the assertion we will now show
that for every ordinal $\alpha$, 
$(a_A^\alpha,b_A^\alpha)\leqi (a_B^\alpha, b_B^\alpha)$.

Clearly, $(a_A^0,b_A^0)\leqi (a_B^0,b_B^0)$. Let us assume that
$\alpha=\beta+1$ and that $(a_A^{\beta},b_A^{\beta})\leqi
(a_B^{\beta}, b_B^{\beta})$. Since $(a_A^{\beta},b_A^{\beta})$ is
$A$-prudent, Lemma \ref{Lem.more.precise.revision} entails that it is
$B$-prudent and 
\[
(a_A^\alpha,b_A^\alpha) =
(\GPL{A}{(b_A^{\beta})},\LO{A}{(a_A^{\beta})}) \leqi
(\GPL{B}{(b_A^{\beta})},\LO{B}{(a_A^{\beta})}).
\]
By Proposition \ref{Prop.monotonicity}, 
\[
(\GPL{B}{(b_A^{\beta})},\LO{B}{(a_A^{\beta})}) \leqi
(\GPL{B}{(b_B^{\beta})}, \LO{B}{(a_B^{\beta})}) = (a_B^\alpha,
b_B^\alpha).
\]

The case of the limit ordinal $\alpha$ is straightforward.

Since $\WF(A)$ and $\WF(B)$ are the limits of the sequences
$\{(a_A^\alpha,b_A^\alpha)\}_{\alpha}$ and $\{(a_B^\alpha,
b_B^\alpha)\}_{\alpha}$, respectively, the second part of the
assertion follows. \hfill$\Box$ \medskip

The next result shows that as the precision of an approximation grows,
all exact fixpoints and exact stable fixpoints are preserved.

\begin{theorem} \label{TST}
Let $O$ be an operator on a complete lattice $L$. Let $A, B \in 
\Appx(O)$. If  $A\leqi B$ then every exact fixpoint of $A$ is an
exact fixpoint of $B$, and every exact stable fixpoint of $A$ 
(that is, an $A$-stable fixpoint of $O$) is also an exact stable
fixpoint of $B$ (that is, a $B$-stable fixpoint of $O$).
\end{theorem}
Proof: Since for every $x\in L$, $A(x,x)=B(x,x)=(O(x),O(x))$, the 
first part of the assertion
follows. Let us now assume that $(x,x)$ is an exact stable fixpoint of
$A$.  In particular, it follows that $(x,x)$ is a fixpoint of $A$ and
is $A$-prudent. By Lemma \ref{Lem.more.precise.revision}, $(x,x)$ is
$B$-prudent and $(x,x)\leqi (\GPL{B}{x},\LO{B}{x})$. The latter pair
is consistent (Proposition \ref{pru1}). Consequently, $(x,x)$ is
$(\GPL{B}{x},\LO{B}{x})$ and hence $x$ is an exact stable fixpoint of
$B$.  \hfill$\Box$ \medskip

Non-exact fixpoints are not preserved, in general. Let us consider
two partial approximations $A$ and $B$ such that $A\leqi B$.
Let us also assume that $\WF(A) <_p \WF(B)$ (that is, $A$ has a
strictly less precise well-founded fixpoint than $B$). Then, clearly,
$\WF(A)$ is no longer a stable fixpoint of $B$. Thus, fixpoints of
$A$ may disappear when we move on to a more precise approximation
$B$.

More precise approximations of a non-monotone operator $O$ yield more
precise well-founded fixpoints and additional exact stable
fixpoints. The natural question is whether there exists an {\em
ultimate approximation} of $O$, that is, a partial approximation most 
precise with respect to the ordering $\leqi$. Such approximation would
have a most precise Kripke-Kleene and well-founded fixpoint and a
largest set of exact stable fixpoints. We will show that the answer to 
this key question is positive. Such ultimate approximation, being 
a distinguished object in the collection of all approximations can be
viewed as determined by $O$. Consequently, fixpoints of the ultimate
approximation of $O$ (including stable, Kripke-Kleene and well-founded 
fixpoints) can be regarded as determined by $O$ and can be associated
with it.

We start by providing a non-constructive argument for the existence of
ultimate approximations. Let us note that the set $\Appx(O)$ is 
not empty. Indeed, let us define $A_O(x,y) = (O(x),O(x))$, if $x = y$,
and $A_O(x,y) = (\bot,\top)$, otherwise. It is easy to see that $A_O\in
\Appx(O)$ and that it is the least precise element in $\Appx(O)$.
Next, we observe that $\Appx(O)$ with the ordering $\leqi$ is a
complete lattice, as the set $\Appx(O)$ is closed under the operations 
of taking greatest lower bounds and least upper bounds. It follows that 
$\Appx(O)$ has a greatest element (most precise approximation). We call 
this partial approximation the {\em ultimate approximation} of $O$ and 
denote it by $U_O$. 

We call the Kripke-Kleene and the well-founded fixpoints of $U_O$, 
the {\em ultimate Kripke-Kleene} and the {\em ultimate well-founded 
fixpoint} of $O$. We denote them by $\KK(O)$ and $\WF(O)$, respectively. 
We call a stable fixpoint of $U_O$ an {\em ultimate partial stable 
fixpoint} of $O$. We refer to an {\em exact} stable fixpoint of
$U_O$ as an {\em ultimate stable fixpoint} of $O$. Exact fixpoints of 
all partial approximations are the same and correspond to fixpoints of 
$O$. Thus, there is no need to introduce the concept of an ultimate 
exact fixpoint of $O$. We have the following corollary to Theorems 
\ref{TKK} and \ref{TST}.

\begin{corollary}\label{mostprecise}
Let $O$ be an operator on a complete lattice $L$. For every $A\in 
\Appx(O)$, $\KK(A) \leqi \KK(U_O)$, $\WF(A)\leqi \WF(U_O)$ and
every $A$-stable fixpoint of $O$ is an ultimate stable
fixpoint of $O$.
\end{corollary}

We will now provide a constructive
characterization of the notion. To state the result, for every $x,y\in L$
such that $x\leq y$, we define $O([x,y])=\{O(z)\colon z\in [x,y]\}$.

\begin{theorem} \label{Segment}
Let $O$ be an operator on a complete lattice $L$. 
Then, for every $(x,y)\in \LC$, $U_O(x,y)=
(\glb(O([x,y])), \lub(O([x,y])))$.
\end{theorem}
Proof: We define an operator $C: L^c \rightarrow L^2$ by setting
\[
C(x,y) = (\glb(O([x,y])), \lub(O([x,y]))).
\]
First, let us notice that since $ \glb(O([x,y])) \leq
\lub(O([x,y]))$, the operator $C$ maps $L^c$ into $L^c$. Moreover, it
is easy to see that $C$ is $\leqi$-monotone. Lastly, since
$O([x,x]) = \{O(x)\}$, 
\[
\glb(O([x,x])) =
\lub(O([x,x])) = O(x).
\]
and, consequently, $C(x,x) = (O(x), O(x))$. Thus, it follows that $C$ 
is a partial approximation of $O$. Since $U_O$ is the most precise 
approximation, we have $C \leq_p U_O$.

On the other hand, $U_O(x,y) \leq (O(z),O(z))$ for every $z \in [x,y]$.
Therefore $U_O^1(x,y) \leqi O(z)$ for all $z \in [x,y]$ and thus
$U_O^1(x,y) \leq \glb(O([x,y]))$. Similarly, $\lub(O([x,y])) \leq
U_O^2(x,y)$. Since $x\leq y$ are arbitrary, $U_O \leq_p C$, as 
desired. $\hfill\Box$

With this result we obtain an explicit characterization of ultimate
stable fixpoints of an operator $O$.
   
\begin{corollary}
Let $L$ be a complete lattice.
An element $x\in L$ is an ultimate stable fixpoint of an operator $O:L\ra 
L$ if and only if $x$ is the least fixpoint of the operator 
$\glb(O([\cdot,x]))$ regarded as an operator on $[\bot,x]$. 
\end{corollary}


We conclude this section by describing ultimate approximations 
for monotone and antimonotone operators on $L$.

\begin{proposition}\label{PropTrivialIsUltimate}
If $O$ is a monotone operator on a complete lattice
$L$ then for every $(x,y)\in \LC$,
$U_O(x,y) = (O(x),O(y))$. If $O$ is antimonotone then for every
$(x,y)\in \LC$, $U_O(x,y) = (O(y),O(x))$.
\end{proposition}
Proof: By Theorem \ref{Segment},
\[
U_O(x,y) = (\glb(O([x,y])), \lub(O([x,y]))).
\]
Now, it is easy to see
that if $O$ is monotone, then $\glb(O([x,y])) = O(x)$ and
$\lub(O([x,y])) = O(y)$. If $O$ is antimonotone, then $\glb(O([x,y]))
= O(y)$ and $\lub(O([x,y])) = O(x)$. The proposition follows.$\hfill\Box$

%

Using the results from \cite{dmt00a} and Proposition
\ref{PropTrivialIsUltimate} we now obtain the following corollary.

\begin{corollary}\label{CorMonAntiMon}
Let $O$ be an operator on a complete lattice $L$. If $O$ is monotone, 
then the least 
fixpoint of $O$ is the ultimate well-founded fixpoint of $O$ and the 
unique ultimate stable fixpoint of $O$. If $O$ is antimonotone, 
then $\KK(O)=\WF(O)$ and every fixpoint of $O$ is an ultimate stable 
fixpoint of $O$. 
\end{corollary}

\section{ULTIMATE SEMANTICS FOR LOGIC PROGRAMMING}

The basic operator in logic programming is the one-step provability
operator $T_P$ introduced in \cite{vEK76}. It is defined on the lattice
of all interpretations. This lattice consists of subsets of the set of
all atoms appearing in $P$ and is ordered by inclusion
(we identify
truth assignments with subsets of atoms that are assigned the value
$\Tr$).

Let $P$ be a logic program. We denote by $U_P$ the ultimate 
approximation operator for the operator $T_P$. By specializing 
Theorem \ref{Segment} to the operator $T_P$ we obtain that for every two
interpretations $I\subseteq J$, 
\[
U_P(I,J) = (\glb(T_P([I,J])), \lub(T_P([I,J]))).
\]
Replacing the ultimate approximation operator $U_O$ in the definitions
of ultimate Kripke-Kleene, well-founded and stable fixpoints with $U_P$
results in the corresponding notions of ultimate Kripke-Kleene,
well-founded and stable models (semantics) of a program $P$. 

We are now in a position to discuss commonsense reasoning intuitions 
underlying abstract algebraic concepts of ultimate approximation and 
its fixpoints. Let us consider two interpretations $I$ and $J$ such 
that $I\subseteq J$. We interpret $I$ as a current lower bound and $J$ 
as a current upper bound on the set of atoms that are true (under $P$). 
Thus, $I$ specifies atoms that are definitely true, while $J$ specifies 
atoms that are possibly true.  Arguably, if an atom $p$ is derived by 
applying the operator $T_P$ to {\em every} interpretation $K\in [I,J]$, 
it can safely be assumed to be true (in the context of the knowledge 
represented by $I$ and $J$). Thus, the set $I'= \glb(T_P([I,J]))$ can 
be viewed as a revision of $I$.

Similarly, since every interpretation $K\in [I,J]$ must be regarded as
possible according to the pair $(I,J)$ of conservative and liberal
estimates, an atom might possibly be true if it can be derived by the
operator $T_P$ from {\em at least} one interpretation in $[I,J]$. 
Thus, the set $J' = \lub(T_P([I,J]))$, consisting of all such atoms, 
can be regarded as a revision of $J$. Clearly, $(I',J')=U_P(I,J)$ 
and, consequently, $U_P$ can be viewed as a way to revise our knowledge 
about the logical values of atoms as determined by a program $P$ from
$(I,J)$ to $(I',J')$. 

By iterating $U_P$ starting at $(\bot,\top)$, we obtain the ultimate 
Kripke-Kleene model of $P$ as an approximation that cannot be further 
improved by applying $U_P$. The ultimate Kripke Kleene model of $P$ 
approximates all fixpoints of $U_P$ and, in particular, all supported
models of $P$. Often, however, the Kripke-Kleene
model is too weak as we are commonly interested in those (partial)
models of $P$ that satisfy some minimality or groundedness conditions. 
These requirements are satisfied by ultimate stable models and, in
particular, by the ultimate well-founded model of $P$. 

When constructing the ultimate well-founded model, we start by assuming 
no knowledge about the status of atoms: no atom is known true and all atoms 
are assumed possible. Our goal is to improve on these bounds.

To improve on the lower bound, we proceed as follows. Our current
knowledge does not preclude any interpretation and all of them
(the whole segment $[\bot,\top]$) need to be taken
into account.  If some atom $p$ can be derived by applying the 
operator $T_P$ to each element of $[\bot,\top]$ then, arguably, $p$ 
could be accepted as definitely true. The set of all these atoms is 
exactly $\glb(T_P([\bot,\top]))$. So, this set, say $I_1$, can be 
taken as a safe new lower bound, giving a smaller interval $[I_1,\top]$
of possible interpretations. We now repeat the same process and obtain 
a new lower bound, say $I_2$, consisting of those atoms that can be 
derived from every interpretation in $[I_1,\top]$. It is given by
$I_2 = \glb(T_P([I_1,\top]))$. Clearly, $I_2$ improves on $I_1$. We 
iterate this process until a fixpoint is reached. This fixpoint, say
$I^1$, consists of all these atoms for which there is a {\em 
constructive} argument that they are true, given that no atoms are
known to be false (all atoms are possible). Thus, it provides a safe 
lower bound for the set of atoms the program should specify as true.

The reasoning for revising the upper bound is different. The goal is
to make false all atoms for which there cannot be a constructive
argument that they are true. Let us consider an interpretation $J$
such that for every $K\in [\bot,J]$, $T_P(K)\in [\bot,J]$, or
equivalently, $lub(T_P([\bot,J]))\subseteq J$. An atom $p \notin J$ 
(false in $J$) cannot be made true by applying $T_P$ to any element
in the segment $[\bot,J]$. In order to derive $p$ by means of $T_P$,
some atoms that are false in $J$ would have to be made true. That,
however, would mean that $p$ is not grounded and could be assumed 
to be false. Thus, each such interpretation $J$ represents an upper
estimate on what is possible (its complement gives a lower 
estimate on what is false) under the assumption that no atom is known 
to be true yet. It turns out that there is a least interpretation, 
say $J^1$ such that $lub(T_P([\bot,J^1]))\subseteq J^1$ and it can 
be constructed in a bottom up way by iterating the operator 
$\lub(T_P([\bot,\cdot])$. This interpretation can be taken as a safe 
lower bound on what is false (given that no atom is known to be true).

The pair $(I^1,J^1)$ is the first improvement on $(\bot,\top)$. It is
precisely the pair produced by the first iteration of the general
well-founded fixpoint definition given earlier. It can now be used, in
place of $(\bot,\top)$, to obtain an even more refined estimate,
$(I^2,J^2)$ and the process continues until the fixpoint is reached.
The resulting pair is the ultimate well-founded model of $P$. This
discussion demonstrates that abstract algebraic concepts of ultimate
approximations can be given a sound intuitive account. 

We will now discuss the properties of the ultimate semantics for logic
programs.

\begin{theorem}
Let $P$, $P'$ be two programs such that $T_P = T_{P'}$. Then, 
the ultimate well-founded models and ultimate stable models of $P$ and
$P'$ coincide.
\end{theorem}
Proof: Theorem \ref{Segment} implies that $U_{P} = U_{P'}$. But
then all fixpoints of $U_{P}$ and $U_{P'}$ coincide. Thus, the
result follows. $\hfill\Box$

This assertion does not hold for the (standard) well-founded and stable
models. For instance, let $P_1 = \{p\lar p,\ \ p\lar \neg p\}$ and $P_2= 
\{p\lar \}$. Clearly, $T_{P_1}=T_{P_2}$. However, $P_2$ has a stable 
model, $\{p\}$, while $P_1$ has no stable models. Furthermore, $p$ is 
true in the well-founded model of $P_2$ and unknown in the well-founded 
model of $P_1$. 

Another appealing property is that the ultimate well-founded model
of a program $P$ with monotone operator $T_P$ is the least fixpoint of
this operator (the least model of $P$). This is a corollary of 
Proposition \ref{CorMonAntiMon}. 
It is not satisfied by the standard well-founded semantics,
as shown by the program $P_1$.

In many cases, the ultimate well-founded semantics coincides with the
standard well-founded semantics. A consequence of Corollary
\ref{mostprecise} is that if the well-founded model of a program is
two-valued, then it coincides with the ultimate well-founded
model. Thus, we have the following result dealing with the classes of
Horn and weakly stratified programs \cite{przy90}:
\begin{proposition}
If a logic program $P$ is a Horn program or a (weakly) stratified
program, then its ultimate well-founded semantics coincides with the 
standard well-founded semantics.
\end{proposition}
Proof: Let $P$ be a Horn program or a weakly stratified program (the
argument is the same). Let $\mathrm{WF}_P$ be the well-founded model 
of $P$. Let $T_P$ be the van Emden-Kowalski operator for $P$, and 
let ${\cal T}_P$ be the corresponding 3-valued operator \cite{fi85}.
Then, ${\cal T}_P$ is an approximation of $T_P$ and the well-founded
model of $P$ satisfies $\mathrm{WF}_P = \WF{({\cal T}_P)}$ 
\cite{dmt00a}. Moreover, for weakly stratified programs, 
$\mathrm{WF}_P$ is two-valued \cite{vrs91}. By Corollary 
\ref{mostprecise}
\[
\mathrm{WF}_P = \WF{({\cal T}_P)} \leq_p \WF{(U_P)}.
\]
Since $\WF{(U_P)}$ is consistent, and $\mathrm{WF}_P$ is complete, it
follows that $\mathrm{WF}_P = \WF{(U_P)}$, as required. $\hfill\Box$

We now show that in general, attractive properties of
ultimate semantics come at a price. 
Namely, we have the following two theorems.

\begin{theorem}
\label{com1}
The problem ``given a finite
propositional logic program $P$, decide whether $P$ has a complete
ultimate stable model'' is $\Sigma^P_2$-complete.
\end{theorem}

\begin{theorem}
\label{com2}
The problems ``given a finite
propositional logic program, compute the ultimate well-founded
fixpoint of $P$'' and ``given a finite propositional logic program,
compute the ultimate Kripke-Kleene fixpoint of $P$'' are in the class
$\Delta^P_2$.
\end{theorem}

These results might put in doubt the usefulness of ultimate semantics.
However, for wide classes of programs the complexity does not grow.
Let $k$ be a fixed integer. We define the class ${\cal E}_k$
to consist of all logic programs $P$ such that for every atom $p\in 
\At(P)$ at least one of the following conditions holds: 
\newcounter{ct35}
\begin{list}{\arabic{ct35}.\ }{\usecounter{ct35}\topsep 0.03in
\parsep 0in\itemsep 0in\labelsep 0in}
\item $P$ contains at most $k$ clauses with $p$ as the head; 
\item the body of each clause with the head $p$ consists of
at most two elements; 
\item the body of each clause with the head $p$ 
contains at most one positive literal; 
\item the body of each clause with the 
head $p$ contains at most one negative literal.
\end{list}

\begin{theorem}
\label{com3}
The problem ``given a finite propositional
logic program from class ${\cal E}_k$, decide whether $P$ has a complete
ultimate stable model'' is {\rm NP}-complete.
\end{theorem}

\begin{theorem}
\label{com4}
The problem ``given a finite
propositional logic program from class ${\cal E}_k$, compute
the ultimate well-founded fixpoint of $P$'' is in {\rm P}.
\end{theorem}

We will now prove these results. If $P$ is a finite propositional
program, then it follows directly from the definition of the ultimate 
Kripke-Kleene fixpoint of $T_P$ (that is, the ultimate Kripke-Kleene model 
of $P$) that it can be computed by means of polynomially many (in the 
size of $P$) evaluations of the operator $U_P(I,J)$, where $I\subseteq 
J$ are interpretations, with all other computational tasks taking only 
polynomial amount of time.

Let us also note that $I$ is a complete ultimate stable model of $P$ 
if and only if $I = \lfp(U_P(\cdot,I))$. Thus, to verify whether $I$ 
is a complete ultimate stable model, it is enough to iterate the 
operator $\lfp(U_P(\cdot,I))$ starting with the empty interpretation. 
The number of iterations needed to reach the least fixpoint is again
polynomial in the size of $P$ with all other needed tasks taking
polynomial time only. A similar discussion shows that the ultimate
well-founded model of $P$ can be computed by means of polynomially
many evaluations of the form $U_P(I,J)$.

It follows that evaluating $U_P(I,J)$, where $I\subseteq J$, is at the 
heart of computing the ultimate Kripke-Kleene, well-founded and
complete stable models of a program $P$. Hence, we will now focus on 
this task. 

Let $P$ be a logic program and let $p$ be an atom in $P$. For every rule 
$r\in P$ such that $p$ is the head of $r$, we define $B_r$ to be the 
conjunction of all literals in the body of $r$. For every atom $p$, we 
denote by $B_P(p)$ the disjunction of all formulas $B_r$, where $r$ 
ranges over all rules in $P$ with the head $p$. When $p$ is the head of 
no rule in $P$ then we set $B_P(r) = \bot$ (empty disjunction). 

Every logic program $P$ has a {\em normal} representation. It
is the collection of rules $p\lar B_P(p)$, where $p$ ranges
over all atoms of $P$. The definition of the operator $T_P$ extends,
in a straightforward way, to the case when $P$ is given in its normal
form defined above.  Moreover, if $P$ is a logic program and $Q$ is
its normal representation, $T_P=T_Q$. Thus, in the remainder of this
section, without loss of generality we will assume that programs are
given by means of their normal representations.

Let us recall that 
\[
U^1_P(I,J)= \glb(T_P([I,J])) = \bigcap_{I\subseteq K\subseteq J}T_P(K)
\]
and 
\[
U^2_P(I,J)= \lub(T_P([I,J])) = \bigcup_{I\subseteq K\subseteq J}T_P(K).
\]
Let $I$ and $J$ be two interpretations such that $I\subseteq J$.  We
define the {\em reduct} $P_{I,J}$ of $P$ to be the program obtained
from $P$ by substituting in each body formula $B_P(p)$, any atom $r$
by ${\bf f}$ if $r\notin J$ and any atom $r$ by ${\bf t}$ if $r\in I$.
Note that all body atoms of $P_{I,J}$ is an element of $J\setminus I$.

We have the following simple properties. An atom $p$ of $P$ belongs to
$U_P^1(I,J)$ if and only if for every interpretation $K\in
[\emptyset,J\setminus I]$, the formula $B_{P_{I,J}}(p)$ is true in $K$
(or, equivalently, if and only if the formula $B_{P_{I,J}}(p)$ is a
tautology). An atom $p$ of $P$ belongs to $U_P^2(I,J)$ if and only if
for some interpretation $K\in [\emptyset,J\setminus I]$, the formula
$B_{P_{I,J}}(p)$ is true in $K$ (or, equivalently, if and only if the
formula $B_{P_{I,J}}(p)$ is satisfiable).

From the second property it follows that computing
$U^2_P(I,J)$ is easy --- it can be accomplished in polynomial
time (in the size of $P$). Indeed, since $B_{P_{I,J}}(p)$ is a DNF
formula, its satisfiability can be decided in polynomial time and the
claim follows. Thus, from now on we will focus on the task of computing 
$U^1_P(I,J)$.  

The problem to decide whether a DNF formula is a tautology is
co-NP-complete. Thus, the problem to compute the ultimate
Kripke-Kleene and well-founded models of a program $P$ is in the 
class $\Delta^P_2$. Consequently, Theorem \ref{com2} follows. 

It also follows that checking whether for an interpretation $J$,
$J=\lfp(U_P^1(\cdot,J))$ is in $\Delta^P_2$. Hence, the problem to decide 
whether a program has a complete ultimate stable fixpoint is in the 
class $\Sigma^P_2$.

We will now show the $\Sigma^P_2$-hardness of the problem of existence 
of a complete ultimate stable model of a program $P$. Let $\vph$ be a
propositional formula and let $I$ be an interpretation (a set of
atoms). We recall that the following problem is $\Sigma^P_2$-complete: 
Given a DNF formula $\vph$ over variables $x_1,\ldots,x_m$, 
$y_1,\ldots, y_n$, decide whether there is a truth assignment 
$I\subseteq \{x_1,\ldots,x_m\}$ such that $\vph_I$ is a tautology,
where $\vph_I$ is the formula obtained by replacing in $\vph$ all 
occurrences of atoms from $I$ with $\Tr$, and by replacing all
occurrences of atoms from $\{x_1,\ldots,x_m\}\setminus I$ with $\Fa$.

We will reduce this problem to our problem. For each $x_i$, $i=1,\ldots,
m$, in $\vph$ we introduce a new variable $x_i'$. We also introduce two
new atoms $p$ and $q$. By $\vph'$ we denote the formula obtained from
$\vph$ by replacing literals $\neg x_i$ in the disjuncts of $\vph$ with 
new atoms $x_i'$. We define a program $P(\vph)$ to consist of the 
following clauses:
\newcounter{ct36}
\begin{list}{\arabic{ct36}.\ }{\usecounter{ct36}\topsep 0.03in
\parsep 0in\itemsep 0in\labelsep 0in}
\item $x_i\lar \n(x_i')$ and $x_i'\lar\n(x_i)$, for every
$i=1,\ldots,m$
\item $y_i\lar \vph'$, for every $i=1,\ldots, n$
\item $p\lar \vph'$
\item $q\lar\n(p),\n(q)$.
\end{list}

We will show that there is $I\subseteq \{x_1,\ldots,x_m\}$ such
that $\vph_I$ is a tautology if and only if $P(\vph)$ has an ultimate
complete stable model.

It is easy to see the that the following properties hold for every 
fixpoint $M$ of $T_{P(\vph)}$: 
\newcounter{ct337}
\begin{list}{\arabic{ct337}.\ }{\usecounter{ct337}\topsep 0.03in
\parsep 0in\itemsep 0in\labelsep 0in}
\item $q$ is false in $M$ (if $q$ is true in $M$, $T_{P(\vph)}$ does
  not derive $q$);
\item $p$ is true in $M$ (otherwise $T_{P(\vph)}$ derives $q$);
\item $y_1,..,y_n$ are true in $M$ (since their rules have the same
bodies as $p$);
\item for each $x_i$, either $x_i$ or $x_i'$ is true in $M$.
\end{list}
For a subset $I\subseteq \{x_1,\ldots,x_m\}$, let us define
$\overline{I} = I\cup \{x_i' : x_i\notin I\}$. It follows from the
properties listed above that for each fixpoint $M$ of $T_{P(\vph)}$ 
and, a fortiori, if $M$ is a complete ultimate stable model of $P(\vph)$,
there exists an $I$ such that
\[
M=\overline{I}\cup \{p,y_1,\ldots,y_n\}.
\]
Thus, it suffices to show that if $I\subseteq \{x_1,\ldots,x_m\}$ then
$M = \overline{I}\cup \{p,y_1,\ldots,y_n\}$ is a complete ultimate
stable model of $P(\vph)$ if and only if $\vph_I$ is a tautology.

It is easy to verify that for every set $M= \overline{I}\cup
\{p,y_1,\ldots,y_n\}$ and for every $J\subseteq M$, $U^1_{}(J,M)$ 
satisfies the following properties:
\newcounter{ct347}
\begin{list}{\arabic{ct347}.\ }{\usecounter{ct347}\topsep 0.03in
\parsep 0in\itemsep 0in\labelsep 0in}
\item $U^1_{P(\vph)}(J,M)\cap \{x_1,..,x_n, x_1',..,x_n'\} =
\overline{I}$ 
\item $U^1_{P(\vph)}(J,M)\cap \{y_1,..,y_n,p,q\}$ is either $\emptyset$
or
$\{y_1,..,y_n,p\}$, since bodies of rules of $y_1,..,y_n,p$ are
identical.
\end{list}
Thus, we find that $U^1_{P(\vph)}(J,M)$ is either $\overline{I}$ or
$M$ and, consequently, 
$U^1_{P(\vph)}(\cdot,M)$ has a least fixpoint, which is either
$\overline{I}$ or $M$. Hence $M= \overline{I}\cup
\{p,y_1,\ldots,y_n\}$ is a complete ultimate stable model of $P(\vph)$
if and only if $\overline{I}$ is not a fixpoint of
$U^1_{P(\vph)}(\cdot,M)$, that is if
$U^1_{P(\vph)}(\overline{I},M)=M$. Consequently, all we need to prove
is that $p\in U^1_{P(\vph)}(\overline{I},M)$ if and only if $\vph_I$ is
a tautology.

Let us recall that $p\in U^1_{P(\vph)}(\overline{I},M)$ if and
only if for every interpretation $K\in [\emptyset,M\setminus
\overline{I}]$, the formula $B_{P(\vph)_{\overline{I},M}}(p)$ is true 
in $K$, that is, if and only if the formula
$B_{P(\vph)_{\overline{I},M}}(p)$ is a tautology.
Let us observe that $B_{P(\vph)}(p) =\vph'$. Thus, it is easy to see
that $B_{P(\vph)_{\overline{I},M}}(p)$ is logically equivalent to
$\vph_I$. Consequently, the claim and Theorem \ref{com1} follows.

The problems of interest restricted to programs from the class ${\cal E}_k$
become easier. Let us recall that the decision whether an atom $p\in 
\At(P)$ belongs to $U^1_P(I,J)$ boils down to the decision whether
the formula $B_{P_{I,J}}(p)$ is a tautology. If $P$ is in the class 
${\cal E}_k$, this question can be resolved in polynomial time. Thus, the
ultimate Kripke-Kleene and the well-founded models for programs in 
${\cal E}_k$ can be computed in polynomial time. Thus, Theorem
\ref{com4} follows. 

Similarly, it takes only polynomial time to verify whether an 
interpretation $I$ satisfies $I=\lfp(U^1_P(\cdot, I))$. Thus, the
problem to decide whether a program from ${\cal E}_k$ has a complete
ultimate stable model is in NP.  
To prove completeness, we observe that for purely negative programs:
\newcounter{ct38}
\begin{list}{\arabic{ct38}.\ }{\usecounter{ct38}\topsep 0.03in
\parsep 0in\itemsep 0in\labelsep 0in}
\item there is no difference between complete stable fixpoints and
complete ultimate stable fixpoints
\item purely negative programs are in ${\cal E}_k$
\item the problem of existence of complete stable fixpoints for purely
negative programs is NP-complete.
\end{list}
Thus, Theorem \ref{com3} follows.

\section{CONCLUSIONS AND DISCUSSION}
\label{conc}

We extended our algebraic framework \cite{dmt00a,dmt00} for studying
semantics of nonmonotonic reasoning systems. The main contribution of
this paper is the notion of an ultimate approximation.  We argue that
the Kripke-Kleene, well-founded and stable fixpoints of the ultimate
approximation of an operator $O$ can be regarded as the Kripke-Kleene,
well-founded and stable fixpoints of the operator $O$ itself. In
earlier approaches, to study fixpoints of an operator $O$ one needed
to {\em select} an appropriate approximation operator. There were,
however, no principled, algebraic ways to do so. In the present paper,
we find a {\em distinguished} element in the space of all
approximations and propose this particular approximation (ultimate
approximation) to study fixpoints of $O$.

A striking feature of our approach is the ease with which it can be
applied in any context where semantics emerge as fixpoints of
operators. We applied this approach here in the context of logic
programming and obtained a family of new semantics for logic programs:
the ultimate Kripke-Kleene, the ultimate well-founded and the ultimate
stable-model semantics. These semantics are well motivated and have
attractive properties. First, they are preserved when we modify the
program, as long as the 2-valued provability operator stays the same
(the property that does not hold in general for standard
semantics). Second, the ultimate Kripke-Kleene and the well-founded
semantics are stronger (in general) than their standard counterparts,
yet approximate the collection of all fixpoints of $O$ and the
collection of all stable fixpoints of $O$, respectively. The
disadvantage is that their complexity is higher. But, as we noticed,
for large classes of programs there is actually no loss in efficiency
of computing ultimate semantics.

This approach can also be applied to default and autoepistemic logics
and results in new semantics with appealing epistemological
features\footnote{We will include a more extensive discussion of these
applications in the journal version of the paper.}. It was also recently
used to define a precise semantics for logic programs with aggregates 
\cite{Denecker:ICLP01}.

We end this discussion with comments on a possible broader role of 
the approximation theory. One common concern when designing semantics 
of nonmonotonic logics is to avoid models justified by ungrounded 
or self-supporting (circular)
arguments. The well-founded fixpoints (semantics)
avoid such arguments. Groundedness is also a fundamental feature of 
induction, a constructive way in which humans specify concepts both 
in commonsense reasoning settings and in formal considerations. In 
its simplest form induction relies only on positive information. In 
general, however, it may make references to negative information, too. 
In either form it is a nonmonotonic specification mechanism. As argued 
in \cite{Denecker98c}, the well-founded semantics 
generalizes existing formalizations of induction (for instance, positive 
induction and iterated induction .

\subsubsection*{Acknowledgments}
This material is based upon work supported by the National Science
Foundation under Grants No. 9874764 and 0097278. Any opinions,
findings, and conclusions or recommendations expressed in this material
are those of the authors and do not necessarily reflect the views of
the National Science Foundation.


{

}

\end{document}